\documentclass[10pt,twocolumn,letterpaper]{article}

\usepackage{wacv}
\usepackage{times}
\usepackage{epsfig}
\usepackage{graphicx}
\usepackage{amsmath}
\usepackage{amssymb}
\usepackage{booktabs}
\usepackage{blindtext}
\usepackage{booktabs}
\usepackage{multirow}
\usepackage{makecell}
\usepackage{subfig}
\usepackage{tikz}

\DeclareMathOperator*{\argmin}{argmin} 

\newcommand\copyrighttextfinal{%
	
	\scriptsize\copyright\ 2022 IEEE. Personal use of this material is permitted. Permission from IEEE must be obtained for all other uses, in any current or future media, including reprinting/republishing this material for advertising or promotional purposes, creating new collective works, for resale or redistribution to servers or lists, or reuse of any copyrighted component of this work in other works. DOI: 10.1109/WACV56688.2023.00263.}%
\newcommand\copyrightnotice{%
	
	\begin{tikzpicture}[remember picture,overlay]%
		
		\node[anchor=south,yshift=10pt] at (current page.south) {{\parbox{\dimexpr\textwidth-\fboxsep-\fboxrule\relax}{\copyrighttextfinal}}};%
	\end{tikzpicture}%
	
}

\def\wacvPaperID{103} 

\wacvfinalcopy 

\ifwacvfinal
\usepackage[breaklinks=true,bookmarks=false]{hyperref}
\else
\usepackage[pagebackref=true,breaklinks=true,colorlinks,bookmarks=false]{hyperref}
\fi

\begin{document}

\title{Heatmap-based Out-of-Distribution Detection}

\author{Julia Hornauer\\
Ulm University, Germany\\
{\tt\small julia.hornauer@uni-ulm.de}
\and
Vasileios Belagiannis\\
Friedrich-Alexander-University Erlangen-Nürnberg, Germany\\
{\tt\small vasileios.belagiannis@fau.de}
}

\maketitle
\thispagestyle{empty}

\begin{abstract}
Our work investigates out-of-distribution (OOD) detection as a neural network output explanation problem. We learn a heatmap representation for detecting OOD images while visualizing in- and out-of-distribution image regions at the same time.
Given a trained and fixed classifier, we train a decoder neural network to produce heatmaps with zero response for in-distribution samples and high response heatmaps for OOD samples, based on the classifier features and the class prediction. Our main innovation lies in the heatmap definition for an OOD sample, as the normalized difference from the closest in-distribution sample. The heatmap serves as a margin to distinguish between in- and out-of-distribution samples. Our approach generates the heatmaps not only for OOD detection, but also to indicates in- and out-of-distribution regions of the input image. In our evaluations, our approach mostly outperforms the prior work on fixed classifiers, trained on CIFAR-10, CIFAR-100 and Tiny ImageNet. The code is publicly available at: \url{https://github.com/jhornauer/heatmap_ood}. 
\end{abstract}

\section{Introduction}
\label{sec:intro}
\copyrightnotice
Despite the astonishing performance of deep neural networks on standard recognition datasets~\cite{Krizhevsky2009LearningML,Russakovsky2015ImageNetLS}, they cannot be trusted yet for safety-critical problems mainly because of two reasons. First, they do not necessarily generalize well to data that was not covered by the training distribution. When deep neural networks are exposed to such out-of-distribution (OOD) samples, they often make wrong predictions with high confidence. Second, they mostly lack providing an explanation on their decision that is understood by humans. If the deep neural network is a black-box model and does not have the necessary tools to assess whether the prediction is meaningful, it cannot be used in applications such as automated driving or medical imaging.
Therefore, it is of utmost importance not only to detect OOD samples but also to highlight out-of-distribution regions of the input.

\begin{figure}
    \centering
    \begin{tabular}{ccc}
        \includegraphics[width=0.21\linewidth]{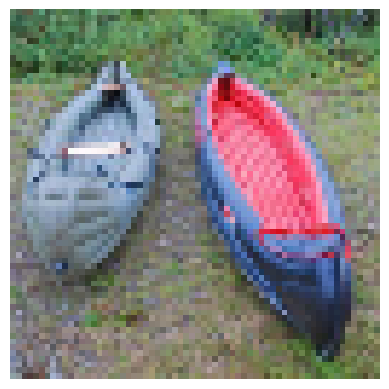} & 
        \includegraphics[width=0.21\linewidth]{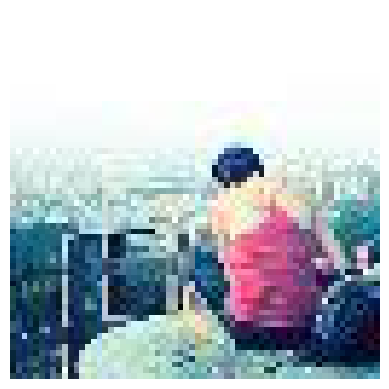} &
        \includegraphics[width=0.21\linewidth]{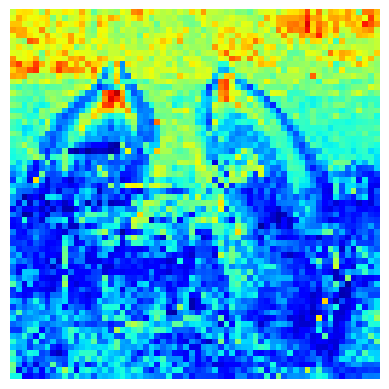} \\
        \subfloat[\label{fig:example_ood}]{ \includegraphics[width=0.21\linewidth]{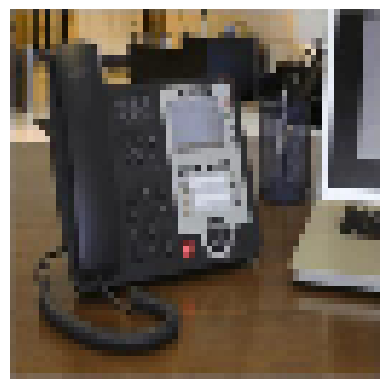}} & 
        \subfloat[\label{fig:example_id}]{ \includegraphics[width=0.21\linewidth]{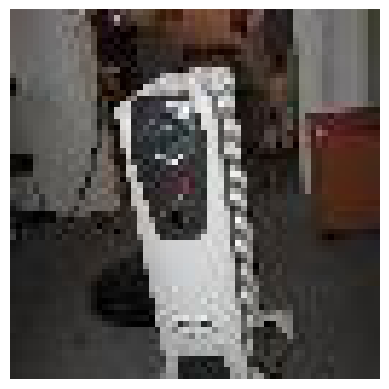}} &
        \subfloat[\label{fig:example_hm}]{ \includegraphics[width=0.21\linewidth]{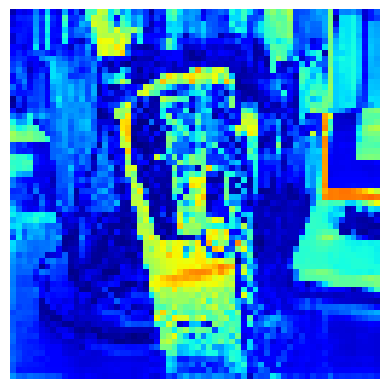}} \\
    \end{tabular}
    \caption{Out-of-distribution training image examples (a) with the corresponding closest in-distribution image (b) and the defined heatmap (c) as a result of the normalized distance between the two images. Blue colors mark similar regions, whereas the red/yellow colors highlight regions that differ from the in-distribution image.}
    \label{fig:examples}
\end{figure}

Out-of-distribution detection methods can be divided into the following two categories: approaches that define a score for a fixed model \cite{hendrycks17_baseline,lee18_mahalanobis} and approaches that further train a model to distinguish in- from out-of-distribution samples \cite{hendrycks2018deep,lee2018training}. In our work, we focus on OOD detection for a fixed classifier, because it suits well to real-world applications.
In this context, the maximum softmax probability~\cite{hendrycks17_baseline} of a neural network often serves as a built-in baseline for the detection of OOD samples. Nevertheless, overconfident deep neural networks usually origin from using the predictive class probability as a confidence measure. The probability distribution represented by the softmax function often gives a high response for unknown inputs~\cite{Nguyen_2015_CVPR}. This limitation can be mitigated by calibrating the confidence~\cite{Liang2018EnhancingTR} or alternative scoring functions~\cite{lee18_mahalanobis,liu20_energy}.
We set ourselves apart by relying on a second model to generate heatmaps, which, in turn, are used to define a scoring function. Thus, our model does not only detect OOD samples but also indicates in- and out-of-distribution image regions.

Our work addresses the detection of OOD samples as an output explanation problem. Approaches towards explaining the model's decision highlight the features that contribute~\cite{Simonyan2014DeepIC,Fukui_2019_CVPR,Selvaraju_2017_ICCV} to the decision or select prototypical images~\cite{barbalau2021generic}. Saliency~\cite{Simonyan2014DeepIC} or attention~\cite{Fukui_2019_CVPR} maps indicate where the trained neural network looks in the image for making a prediction. Prototypes on the other hand, demonstrate a similar training image~\cite{barbalau2021generic} for interpreting the neural network prediction. Similar to post hoc explanation approaches~\cite{barbalau2021generic,Simonyan2014DeepIC,Selvaraju_2017_ICCV}, we also assume a model that is already trained and will not be further adapted. Motivated by the visual explanation idea, we propose the heatmap generation for OOD detection.

We phrase the problem of OOD detection as binary classification in the classifier's feature space assuming that the features extracted by the classifier are representative to learn the deviation of out-of-distribution from in-distribution features. Given a trained and fixed neural network classifier, we aim to generate heatmaps based on the classifier features and the class prediction. Therefore, we define the expected heatmaps such that in-distribution samples should have zero response. In contrast, OOD samples should have a high response for image regions that differ from the in-distribution samples. We create the heatmap of an OOD image by looking for the closest in-distribution image in the classifier feature space and forming the heatmap as the normalized image difference (see Fig.~\ref{fig:examples}). In this way, the heatmap response acts as a margin between the in- and out-of-distribution image samples. To develop our idea, we introduce a decoder network to generate the heatmaps with input the features extracted by the classifier and the class prediction, while the in- and out-of-distribution defined heatmaps compose the expected output. Finally, based on the heatmaps, we define our scoring function, which outputs whether a sample is in- or out-of-distribution. Our approach generates the heatmaps not only for OOD detection but also as visualization of in- and out-of-distribution regions based on the classifier features and the class prediction.  
We summarize our contribution as follows: First, we propose the heatmaps for OOD detection. Learning to generate the heatmaps is based on our introduced decoder, while our main novelty lies in how to create heatmaps for the OOD image samples. Second, the proposed heatmaps visualize in- and out-of-distribution image regions based on the classifier features and the class prediction. In particular, for OOD samples the heatmaps represent the difference to the closest in-distribution sample as illustrated in Fig~\ref{fig:examples}. Third, we present the OOD scoring function using the heatmap as input, which results in state-of-the-art OOD detection results for different classifiers trained on CIFAR-10~\cite{Krizhevsky2009LearningML}, CIFAR-100~\cite{Krizhevsky2009LearningML} or Tiny ImageNet~\cite{Russakovsky2015ImageNetLS}, compared to approaches that do not modify the classifier. 

\section{Related Work}
\label{sec:relatedwork}
\paragraph{Out-of-Distribution Detection}
Out-of-distribution detection approaches aim to identify samples that are not covered by the training distribution \cite{hendrycks17_baseline}. The prior work on image classification is mainly divided into two categories, based on whether the classifier parameters are fixed or modifiable. For fixed deep neural classifiers, the maximum softmax probability (MSP) serves as a common function to detect OOD samples. Nevertheless, the softmax probability is not sufficient due to the major drawback of deep neural networks being overconfident for unseen samples~\cite{Nguyen_2015_CVPR}. ODIN~\cite{Liang2018EnhancingTR} improves the softmax score with temperature scaling and input perturbations validated on OOD samples. Hsu et al.~\cite{Hsu2020GeneralizedOD} extend the temperature scaling to be independent of the OOD validation data. 
Liu et al.~\cite{liu20_energy} introduce the energy score to discriminate between in- and out-of-distribution samples of a pre-trained classifier, but it also can be used as a cost function to optimize the classifier for the detection of OOD samples. 
Lin et al.~\cite{Lin2021MOODMO}, on the other hand, implement the energy score with an early exit strategy to address faster OOD detection. Sun et al.~\cite{Sun2021ReActOD} use the observation that OOD data causes positively skewed activation units and improve the detection of OOD data by clipping the activations of the penultimate layer to an upper limit based on in-distribution activation values. We also assume a fixed classifier that is not further modified. Unlike the prior work~\cite{hendrycks17_baseline,lee18_mahalanobis,Liang2018EnhancingTR}, we learn to separate in-distribution from out-of-distribution samples with the heatmaps, generated by our proposed decoder. Importantly, our decoder not only performs OOD detection but also delivers a visualization of out-of-distribution regions through heatmap responses.
To our knowledge, this is the first approach to perform OOD detection and at the same time illustrate in- and out-of-distribution image regions based on the features and the class prediction of an already trained and fixed classifier. 
In the second category, where the classifier is further optimized to not only classify the network inputs but also determine OOD samples, self-supervised methods such as deep autoencoders \cite{Pidhorskyi2018GenerativePN} or rotation prediction \cite{NEURIPS2019_a2b15837} can be leveraged to learn the in-distribution representation.
Zaeemzadeh et al.~\cite{Zaeemzadeh2021OutofDistributionDU}, for instance, rely on feature compression by learning to embed in-distribution samples to a 1-dimensional subspace for OOD detection. Zisselman et al.~\cite{Zisselman2020DeepRF} design a network architecture based on deep residual flow networks, customized for OOD detection, while Huang et al.~\cite{huang2021mos} target large-scale OOD detection with a group-based softmax. In addition, there are approaches that utilize OOD training samples during the classifier's training stage. For example, random~\cite{hendrycks2018deep} or generated~\cite{lee2018training} OOD samples can be mapped to the Uniform distribution. In contrast, Yang et al.~\cite{yang2021scood}, apply clustering in the semantic space to detect in-distribution data within the OOD training samples. Based on the same principle, we also use OOD samples for the training of the proposed decoder, but basically without modifying the original classifier. We rely on the features learned by the already trained classifier to distinguish out-of-distribution from in-distribution samples. 

\paragraph{Prediction Explainability}
There are different approaches to explain neural network predictions. Post hoc methods give explanations for trained models by local explanations \cite{Ribeiro2016WhySI} or global approximations with an interpretable surrogate model \cite{Lakkaraju2019FaithfulAC}. Another line of research visualizes prototypes \cite{barbalau2021generic} or highlights features that contribute to the classifier decision through activation \cite{Selvaraju_2017_ICCV}, attention \cite{Fukui_2019_CVPR}, or saliency \cite{Simonyan2014DeepIC} maps. 
Moreover, Liznerski et al.~\cite{expldocc_liznerski} make use of visual explanation to highlight conspicuous regions in images for anomaly detection. In this work, we phrase OOD detection as anomalies in the feature representation of an already trained and fixed model.
In this context, we propose to generate heatmaps from the features extracted by a classifier in order to detect out-of-distribution inputs and visualize the corresponding regions of the input image.

\section{Method}
Let $P_{in}(\mathbf{x}, \mathbf{y})$ be the joint distribution of the image $\mathbf{x} \in \mathbb{R}^{w \times h \times 3}$ with width $w$ and height $h$, as well as the one-hot vector label $\mathbf{y} \in \{0, 1\}^{C}$, such that $\sum_{c=1}^{C}\mathbf{y}(c)=1$ for a $C$-category classification problem. In our context, $P_{in}$ denotes the training distribution from which the in-distribution dataset $\mathcal{D}_{in}=\{(\mathbf{x}_{i}, \mathbf{y}_{i})\}_{i=1}^{|\mathcal{D}_{in}|}$ is generated. Moreover, the multi-class deep neural network classifier $f(\cdot)$, parameterized by $\mathbf{w_{f}}$, is trained with the $\mathcal{D}_{in}$ dataset. Importantly, we consider the parameters of $f(\cdot)$ to be fixed and not modifiable anymore. During deployment, the classifier $f(\cdot)$ can be exposed to the data distribution $P_{out}(\mathbf{x}, \mathbf{y})$ that is different from the training distribution. We also consider the OOD training set  $\mathcal{D}_{out}=\{(\mathbf{x}_{o}, \mathbf{y}_{o})\}_{o=1}^{|D_{out}|}$ that is formed by sampling from $P_{out}$. In general, images $\mathbf{x}$ drawn from $P_{out}$ have either a non-semantic shift or belong to a different object category \cite{Hsu2020GeneralizedOD}.

\begin{figure}
    \centering
    \includegraphics[width=\linewidth]{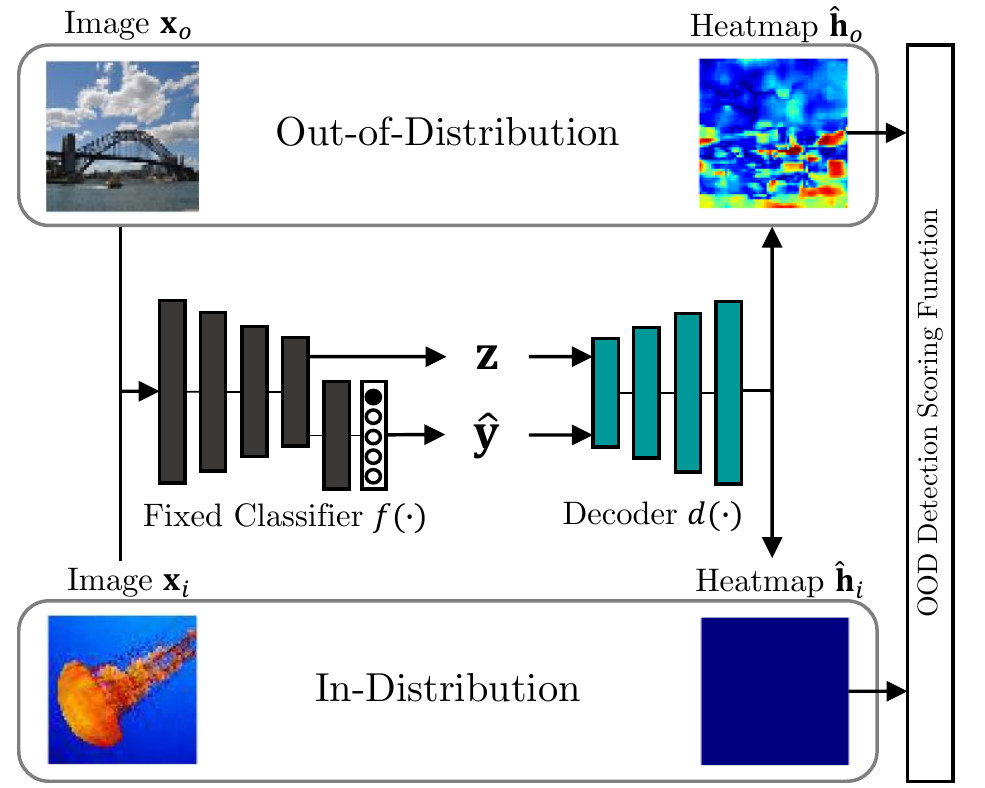}
    \caption{In our approach, we aim to detect OOD samples and at the same time visualize in- and out-of-distribution regions. Therefore, we generate heatmaps from the features $\mathbf{z}$ and the prediction $\mathbf{\hat{y}}$ of a fixed classifier. The generated heatmaps $\mathbf{\hat{h}}_{i}$ should show no response for in-distribution images $\mathbf{x}_{i}$, while the heatmaps $\mathbf{\hat{h}}_{o}$ should have a high response for OOD images $\mathbf{x}_{o}$.}
    \label{fig:Overview}
\end{figure}

Our goal is to differentiate in-distribution $P_{in}(\mathbf{x}, \mathbf{y})$ from out-of-distribution $P_{out}(\mathbf{x}, \mathbf{y})$ image samples based on the prediction and the features of the classifier $f(\cdot)$.
We formulate the problem as binary classification where we assume that we have access to the feature layers of the fixed classifier $f(\cdot)$. In this context, we introduce the OOD heatmaps, which are utilized for spotting OOD samples and simultaneously illustrate in- and out-of-distribution image regions~(\ref{sec:heatmaps}). 
Our main innovation lies in how to form expected heatmaps for the out-of-distribution samples. Also, we present the decoder neural network $d(\cdot)$, which is trained to generate heatmaps (\ref{sec:training}) from the classifier's features and class prediction. 
Finally, we rely on the decoder's heatmap generation for each image sample to compute the OOD detection score, as discussed in Sec.~\ref{sec:oodscore}. 
Essentially, the generated heatmap indicates the corresponding image regions for the in- or out-of-distribution predictions (see Fig.~\ref{fig:Overview}).
Note that for the evaluation, we consider the OOD test set $\mathcal{D}_{test}=\{(\mathbf{x}_{l}, \mathbf{y}_{l})\}_{l=1}^{|\mathcal{D}_{test}|}$ from a different distribution $P_{test}(\mathbf{x}, \mathbf{y})$ that is not covered by $P_{in}$ or $P_{out}$. 

\subsection{Heatmaps}\label{sec:heatmaps}
We define the expected heatmap $\mathbf{h} \in \mathbb{R}^{w \times h \times 3}$ with the same dimensions as the input image $\mathbf{x}$. For an in-distribution image, the heatmap should have zero response since it does not contain out-of-distribution information. In contrast, the heatmap of an out-of-distribution sample should have a high response for the image regions that differ from the in-distribution data. We provide an illustration of the heatmaps in Fig.~\ref{fig:Overview}. Consequently, the heatmap response acts as a margin between the in- and out-of-distribution image samples.
For the in-distribution images, it is trivial to specify the expected heatmap since we aim to have a zero response. However, it is not clear how to define the heatmap for out-of-distribution samples. Below, we present an approach to make use of the in-distribution dataset $\mathcal{D}_{in}$ for defining the expected heatmaps of the images of the out-of-distribution dataset $\mathcal{D}_{out}$.

\begin{figure}
    \centering
    \includegraphics[width=0.95\linewidth]{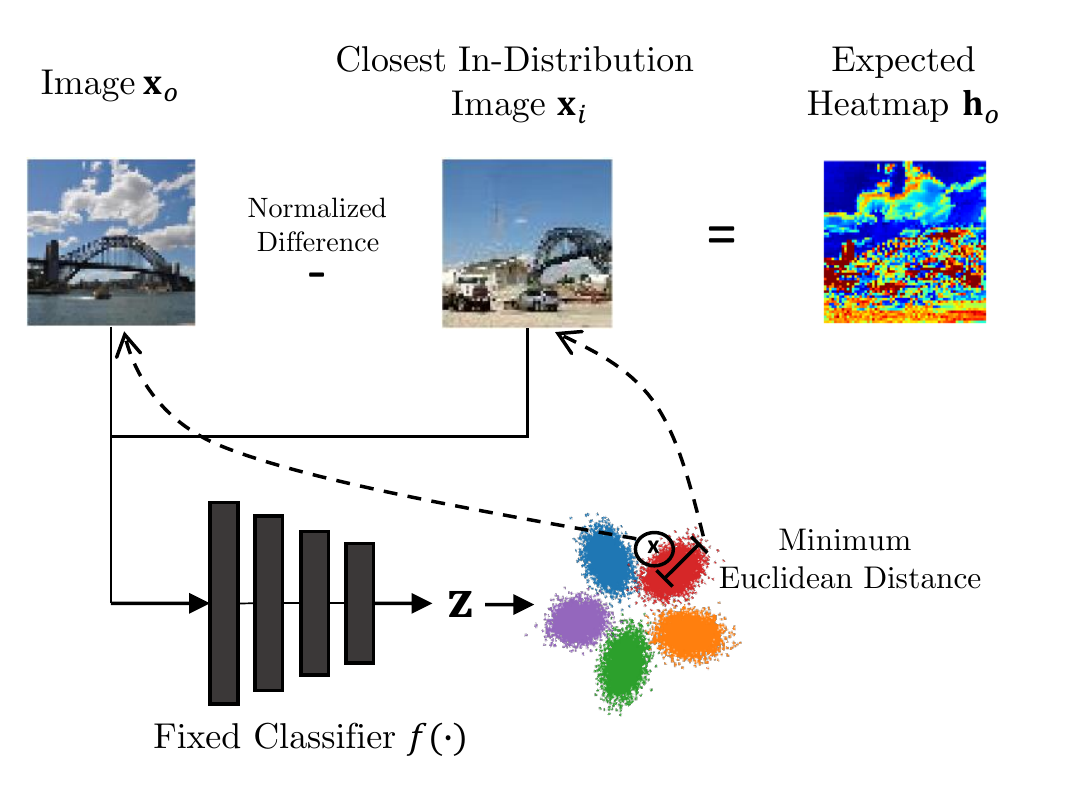}
    \caption{The definition of the expected OOD heatmap $\mathbf{h}_{o}$ is based on the distance to the closest in-distribution sample from $\mathcal{D}_{in}$. To find the closest in-distribution sample, we rely on the class prediction $\mathbf{\hat{y}}$ and the feature representation $\mathbf{z}$ of the fixed classifier. More precisely, we look for the sample $\mathbf{x}_{i}$ from $\mathcal{D}_{in}$ with the same class category, i.e. $\mathbf{\hat{y}}_{o}=\mathbf{y}_{i}$ and minimum Euclidean distance between $\mathbf{z}_{o}$ and $\mathbf{z}_{i}$.}
    \label{fig:HeatmapOOD}
\end{figure}

\paragraph{\textbf{Out-of-Distribution Heatmap.}}
 For each image sample $\mathbf{x}_{o}$ of the out-of-distribution dataset $\mathcal{D}_{out}$, our idea is to form the heatmap based on the closest image sample of the in-distribution dataset $\mathcal{D}_{in}$.
For that reason, we rely on the class prediction $\mathbf{\hat{y}}_{o}$ and the feature representation $\mathbf{z}_{o}$ of the fixed classifier with $\mathbf{\hat{y}}_{o}, \mathbf{z}_{o} = f(\mathbf{x}_{o}; \mathbf{w}_{f})$. The selection is illustrated in Fig.~\ref{fig:HeatmapOOD}. For the feature representation $\mathbf{z}_{o}$, we use the penultimate layer of the classifier.
Based on these two attributes, we look for the sample $\mathbf{x}_{i}$ from $\mathcal{D}_{in}$ with the same class category, i.e.~$\mathbf{\hat{y}}_{o} = \mathbf{y}_{i}$, and similar feature representation, i.e.~$\mathbf{z}_{o} \approx \mathbf{z}_{i}$. Note that even if $\mathbf{\hat{y}}_{o}$ is a wrong prediction, it does not have an impact on our approach. The goal is to find the closest in-distribution sample $\mathbf{x}_{i}$ based on the Euclidean distance between $\mathbf{z}_{o}$ and $\mathbf{z}_{i}$ given the class category. After obtaining the corresponding in-distribution sample $\mathbf{x}_{i}$ for the out-of-distribution sample $\mathbf{x}_{o}$, we define the heatmap $\mathbf{h}_{o}$ as their difference, where the images are normalized beforehand. The operation is described as: 
\begin{equation}
\mathbf{h}_{o} = norm(\mathbf{x}_{i^{*}}) - norm(\mathbf{x}_{o}), 
\end{equation}
\begin{equation}
\begin{gathered} 
\text{where} \quad i^{*}= \argmin_{i \in \{1, \dots ,|\mathcal{D}_{in}|\}} (\mathbf{z}_{i} - \mathbf{z}_{o})^{2}, \\\
     \text{s.t.~} \mathbf{\hat{y}}_{o} = \mathbf{y}_{i}. 
\end{gathered}
\end{equation}
The $norm(\cdot)$ operation corresponds to the normalization operation within the range $-1$ and $1$.
By considering the normalized image difference as the heatmap response, we define a margin between the in- and out-of-distribution image samples. Moreover, the margin is deliberated because both images have similar feature representations. In this way, we create the corresponding heatmap $\mathbf{h}_{o}$ for each image $\mathbf{x}_{o}$ of $\mathcal{D}_{out}$, resulting in our out-of-distribution training set. The complete image and heatmap set is defined as:
\begin{equation}
\mathcal{H}_{out} = \{ (\mathbf{x}_{o}, \mathbf{h}_{o}) \}_{o=1}^{|D_{out}|}.
\end{equation}
We consider $\mathcal{H}_{out}$ as the out-of-distribution training set for training the proposed decoder. 

\paragraph{\textbf{In-Distribution Heatmap.}}
In contrast to out-of-distribution samples, the zero response heatmap $\mathbf{h}_{i}= [0]^{w \times h \times 3}$ is representative for in-distribution images. 
These samples do not contain any out-of-distribution image regions to be indicated in the heatmap. 
Given the in-distribution images $\mathbf{x}_{i}$ and their heatmaps $\mathbf{h}_{i}$, we define our in-distribution training set as:
\begin{equation}
    \mathcal{H}_{in} = \{(\mathbf{x}_{i}, \mathbf{h}_{i})\}_{i=1}^{|\mathcal{D}_{in}|}.
\end{equation}
Based on $\mathcal{H}_{in}$ and $ \mathcal{H}_{out}$, we can train the proposed decoder to generate heatmaps for both types of image samples.

\paragraph{\textbf{Heatmap Interpretation.}}
The visual comparison between in- and out-of-distribution images is not directly possible in the image space due to the pixel value ambiguity and generally complex space. In our approach, the decoder learns a mapping from feature space back to pixel space, where the heatmap represents the difference to the closest in-distribution image. Therefore, OOD detection is based not only on the information in feature space, but also on the mapping back to pixel space learned by the decoder, which improves the OOD detection performance. In Fig.~\ref{fig:examples}, features that differ from the closest in-distribution training sample are highlighted by red regions, which in turn are OOD features. In contrast, the blue color indicates similar features that are in-distribution regions.

\subsection{Heatmap Decoder}\label{sec:training}
The proposed decoder $d(\cdot)$, parameterized by $\mathbf{w_{d}}$, aims to estimate the heatmaps $\mathbf{\hat{h}}$. Similar to the heatmap definition, we rely on the feature representation $\mathbf{z}$ and the predictive probability distribution $\mathbf{\hat{y}}$ as input to the decoder. Fig.~\ref{fig:Overview} shows an overview of our approach. At first, the classifier receives the image $\mathbf{x}$ either drawn from $P_{in}$ or drawn from $P_{out}$ as input and outputs the corresponding feature representations $\mathbf{z}$ and the predictive probability distribution $\mathbf{\hat{y}}$.
Then, the features and the one-hot encoded class prediction are concatenated before being passed to the decoder to estimate the heatmaps $\mathbf{\hat{h}}$. Again, we rely only on a single feature representation of $\mathbf{z}$, which is the penultimate classifier layer. 
In the supplementary material we add an ablation study to validate this choice. 
We use thereby the $\mathcal{H}_{in}$ and $ \mathcal{H}_{out}$ sets to train the decoder producing heatmaps. Below, the learning objective of the decoder is summarized as: 
\begin{equation}
\begin{gathered}
\argmin_{\mathbf{w_{d}}} \mathbb{E}_{\mathbf{x}_{o}, \mathbf{h}_{o} \sim \mathcal{H}_{out}} (1 + \alpha | \mathbf{h}_{o} |) \parallel \mathbf{\hat{h}}_{o} - \mathbf{h}_{o} \parallel^{2} +\\\ \mathbb{E}_{\mathbf{x}_{i}, \mathbf{h}_{i} \sim \mathcal{H}_{in}} \parallel \mathbf{\hat{h}}_{i} -  \mathbf{h}_{i} \parallel^{2},
\end{gathered}
\end{equation}
\begin{equation}
 \text{where} \quad  \mathbf{\hat{h}}_{\{i,o\}} = d(f(\mathbf{x}_{\{i,o\}}; \mathbf{w_{f}}); \mathbf{w_{d}}).
\end{equation}
The OOD entries in the heatmap are weighted higher with a scaling factor $1 + \alpha | \mathbf{h}_{o} |$ depending on the defined heatmap $\mathbf{h}_{o}$. Since in-distribution regions occur more frequently, the scaling factor has the effect that out-of-distribution regions are weighted higher in the loss calculation. 
For the in-distribution samples, we minimize effectively the term $\parallel \mathbf{\hat{h}_{i}} \parallel^{2}$ since the $\mathbf{h}_{i}$ is always zero for in-distribution samples. 
During inference, where we have no knowledge of the distribution, the trained decoder estimates the heatmaps $\mathbf{\hat{h}}$ for illustrating in- and out-of-distribution image regions and for defining our OOD scoring function. 

\subsection{Out-of-Distribution Scoring Function}\label{sec:oodscore}
We leverage the generated heatmap $\mathbf{\hat{h}}$ to define the OOD scoring function for differentiating in- from out-of-distribution images.
Since OOD detection is a binary classification problem, the OOD scoring function is described as:
\begin{equation} \label{ood_score}
    OOD(\mathbf{\hat{h}}) = \begin{cases} 
          1 & \frac{1}{w \cdot h} \sum_{j=1}^{w \cdot h} \max  (|\mathbf{\hat{h}}_{j}|) \leq \delta \\
          0 & \text{otherwise},
       \end{cases}
\end{equation}
where $j$ is used to iterate over the heatmap pixels and $\delta$ is the threshold to categorize a sample as in- or out-of-distribution. 

\section{Experiments}
\label{sec:experiments}
We conduct a detailed evaluation on standard OOD detection benchmarks using different network architectures. Therefore, we compare our method to state-of-the-art OOD detection approaches that operate on a pre-trained model (\ref{sec:oodresults}). We demonstrate the heatmaps that serve as illustration of in- and out-of-distribution image regions (\ref{sec:visu}) based on the distance to the closest  in-distribution training sample. To further explain the heatmaps, we evaluate the effect of lighting conditions by influencing the brightness of images (\ref{sec:br}).
Lastly, we show the importance of the OOD training data size in an ablation study (\ref{sec:ablation}). 

\subsection{Experimental Setting}
\paragraph{\textbf{Datasets and Models.}}
\label{sec:datasets}
We follow the prior work~\cite{hendrycks2018deep,liu20_energy} to evaluate our method on CIAFR-10~\cite{Krizhevsky2009LearningML} and CIFAR-100~\cite{Krizhevsky2009LearningML} as in-distribution datasets. Both in-distribution datasets have a resolution of $32 \times 32$. To cover a wide variety of OOD samples, we rely on the following OOD test sets: iSUN~\cite{Xu2015TurkerGazeCS}, LSUN-Crop~\cite{Yu2015LSUNCO}, LSUN-Resize~\cite{Yu2015LSUNCO}, SVHN~\cite{Netzer2011ReadingDI}, Textures~\cite{Cimpoi2014DescribingTI}, and Places365~\cite{Zhou2018PlacesA1}. All OOD images are downsampled to the in-distribution image resolution. As in the prior work \cite{hendrycks2018deep}, we use the 80 Million Tiny Images~\cite{Torralba200880MT} database as the OOD training set. In this context, we train ResNet18~\cite{He2016DeepRL} and WideResNet~\cite{BMVC2016_87} with depth 40 and width 2 on the in-distribution datasets and fix the classifier weights afterwards.
In addition, we evaluate our method on the complex setting with ResNet50~\cite{He2016DeepRL} trained on Tiny ImageNet~\cite{Russakovsky2015ImageNetLS} as an in-distribution database. Here, the image resolution is $64 \times 64$. For the OOD test set, we rely on the iNaturalist~\cite{Horn2018TheIS}, SUN~\cite{Xiao2010SUNDL}, and Textures~\cite{Cimpoi2014DescribingTI} databases, which are again downsampled to the same resolution as the in-distribution images.
For the complex setup, we leverage Places365 as an OOD training set. As in prior work~\cite{hendrycks2018deep,liu20_energy}, we evaluate with the entire in-distribution test set. As in literature~\cite{hendrycks2018deep,liu20_energy}, the number of OOD samples per OOD test set is fixed to $\frac{1}{5}$ of the in-distribution dataset size.

\paragraph{\textbf{Decoder Architecture and Training.}}\label{sec:implementation}
The decoder network architecture is based on the DCGAN generator~\cite{Radford2016UnsupervisedRL}. The heatmaps are normalized to $[-1, 1]$ with hyperbolic tangent as the last activation function in the decoder. 
As mentioned in \ref{sec:training}, we rely on a single feature representation of the classifier, namely the penultimate classifier layer, as input to the decoder. Then, the features are normalized to the range of $[0, 1]$ and afterward concatenated with the one-hot encoded class prediction. Furthermore, the scaling factor $\alpha$ of the loss function is empirically chosen to be $5$.
Furthermore, the decoder is trained for $150$ epochs with the Adam\cite{Kingma2015AdamAM} solver, where the learning rate is set to $0.0002$, $\beta_{1}$ to $0.5$, and $\beta_{2}$ to $0.999$. Finally, the input images are processed in a batch of $200$ samples. The ratio from OOD samples to in-distribution samples is empirically chosen to be $\frac{1}{5}$. The CIFAR-10 and CIFAR-100 databases both consist of $50000$ training images. This means the number of OOD training samples is set to $10000$. For Tiny ImageNet, which contains $100000$ training images, the OOD training data size then is $20000$. Similar to prior work~\cite{hendrycks2018deep}, the OOD training images are randomly chosen. 

\paragraph{\textbf{Evaluation Metrics.}}
\label{sec:metrics}
We evaluate our method based on the standard metrics~\cite{hendrycks2018deep,liu20_energy,Zaeemzadeh2021OutofDistributionDU}, namely the AUROC, AUPR-Success (AUPR-S), AUPR-Error (AUPR-E), and FPR at 95\% TPR (FPR-95). All metrics are independent of the OOD detection threshold $\delta$. The AUROC integrates over the area under the receiver operating curve (ROC). The AUPR-Success indicates the area under the precision-recall curve with in-distribution samples as positive, while the AUPR-Error treats the OOD samples as positive. Compared to the AUROC metrics, the AUPR accounts for class imbalance. Lastly, FPR at 95\% TPR computes the false positive rate (FPR) at 95\% true positive rate (TPR). We directly apply the metric to the respective OOD detection score. 

\paragraph{\textbf{Comparison to Related Work.}}
\label{sec:comparison}
We compare with post hoc methods that operate on a fixed classifier, similar to our approach. First, we employ the maximum softmax probability (MSP)~\cite{hendrycks17_baseline}. In addition, we use ODIN~\cite{Liang2018EnhancingTR} and Mahalanobis~\cite{lee18_mahalanobis} for comparison with our approach. For ODIN, we set the temperature to $1000$, as proposed in the paper, and select the noise magnitude to be $0.0014$ based on the best outcome. In the case of Mahalanobis, we obtain the best results with a noise magnitude of $0.0028$ when relying only on the features of the penultimate classifier layer. Furthermore, we compare our method to the recent Energy score~\cite{liu20_energy} and ReAct~\cite{Sun2021ReActOD}. 
In case of the Energy score, we choose the version that does not further optimize the classifier. Instead, the score is also determined based on the fixed classifier, since we train a second model but do not adjust the classifier. 
The temperature is set to $1$ as proposed by the authors.
For the related approaches, we make use of the official implementations. All approaches are evaluated on the same network architecture. Since the authors of the Energy score also conduct their evaluation with a pre-trained WideResNet, we additionally report the original results from their paper marked as Energy$^{\ast}$.
Since our approach is based on a trainable model, unlike the related approaches, we conduct each experiment five times and report the mean value over all runs. 

\begin{table*}
    \begin{center}
    \small{
    \begin{tabular}{clcccc}
    \toprule
        $D_{in}$ & \multirow{2}{*}{Method} & \multirow{2}{*}{AUROC $\uparrow$} & \multirow{2}{*}{AUPR-S $\uparrow$} & \multirow{2}{*}{AUPR-E $\uparrow$} & \multirow{2}{*}{FPR-95 $\downarrow$} \\
        (model) & & & & & \\
        \midrule
        \multirow{6}{*}{\makecell{CIFAR-10 \\ (ResNet)}} & MSP \cite{hendrycks17_baseline} & 90.72 & 97.89 & 63.48 & 55.21 \\
        & ODIN \cite{Liang2018EnhancingTR} & 88.33 & 96.67 & 71.49 & 38.35 \\
        & Mahalanobis \cite{lee18_mahalanobis} & 92.33 & 98.29 & 71.30 & 39.52 \\
        & Energy \cite{liu20_energy} & 91.72 & 97.90 & 72.12 & 37.97 \\
        & ReAct \cite{Sun2021ReActOD} & 91.71 & 97.89 & 72.55 & 36.52 \\
        & Ours & \textbf{96.47} & \textbf{99.17} & \textbf{83.73} & \textbf{15.37} \\
        \midrule
        \multirow{7}{*}{\makecell{CIFAR-10 \\ (WideResNet)}} & MSP \cite{hendrycks17_baseline} & 91.48 & 98.18 & 63.47 & 56.77 \\
        & ODIN \cite{Liang2018EnhancingTR} & 95.01 & 98.68 & 84.39 & 21.09 \\
        & Mahalanobis \cite{lee18_mahalanobis} & 92.03 & 98.09 & 75.44 & 32.73 \\
        & Energy \cite{liu20_energy} & 94.91 & 98.75 & 80.89 & 24.26 \\
        & Energy$^{\ast}$ \cite{liu20_energy} & 91.88 & 97.83 & - & 33.01 \\
        & ReAct \cite{Sun2021ReActOD} & 51.92 & 85.46 & 17.53 & 97.12 \\
        & Ours & \textbf{96.36} & \textbf{99.07} & \textbf{86.73} & \textbf{14.06} \\
        \midrule
        \multirow{6}{*}{\makecell{CIFAR-100 \\ (ResNet)}} & MSP \cite{hendrycks17_baseline} & 79.29 & 95.04 & 40.34 & 76.58 \\
        & ODIN \cite{Liang2018EnhancingTR} & 83.28 & 95.96 & 48.74 & 67.96 \\
        & Mahalanobis \cite{lee18_mahalanobis} & 73.46 & 93.00 & 35.90 & 79.46 \\
        & Energy \cite{liu20_energy} & 82.07 & 95.71 & 43.92 & 74.45 \\
        & ReAct \cite{Sun2021ReActOD} & 84.22 & 96.27 & 49.08 & 67.78 \\
        & Ours & \textbf{86.74} & \textbf{96.49} & \textbf{58.78} & \textbf{52.73} \\
        \midrule
        \multirow{7}{*}{\makecell{CIFAR-100 \\ (WideResNet)}} & MSP \cite{hendrycks17_baseline} & 65.31 & 90.38 & 26.21 & 88.45 \\
        & ODIN \cite{Liang2018EnhancingTR} & 79.43 & 94.60 & 43.98 & 73.19 \\
        & Mahalanobis \cite{lee18_mahalanobis} & 73.99 & 92.58 & 43.80 & 68.45 \\
        & Energy \cite{liu20_energy} & 77.11 & 93.95 & 39.07 & 78.03 \\
        & Energy$^{\ast}$ \cite{liu20_energy} & 79.56 & 94.87 & - & 73.60 \\
        & ReAct \cite{Sun2021ReActOD} & 80.74 & 95.24 & 48.04 & 67.47 \\
        & Ours & \textbf{85.98} & \textbf{95.96} & \textbf{61.14} & \textbf{49.86} \\  
        \midrule
        \multirow{6}{*}{\makecell{Tiny ImageNet \\ (ResNet)}} & MSP \cite{hendrycks17_baseline} & 72.16 & 93.12 & 29.06 & 87.93 \\
        & ODIN \cite{Liang2018EnhancingTR} & 75.25 & 94.01 & 32.59 & 85.67 \\
        & Mahalanobis \cite{lee18_mahalanobis} & 74.99 & 93.01 & 44.03 & 68.97 \\
        & Energy \cite{liu20_energy} & 75.99 & 94.20 & 33.74 & 84.00 \\
        & ReAct \cite{Sun2021ReActOD} & \textbf{85.53} & \textbf{96.50} & 54.52 & 61.10 \\
        & Ours & 85.28 & 96.25 & \textbf{56.14} & \textbf{54.66} \\
    \bottomrule
    \end{tabular}}
    \end{center}
    \caption{Comparison of the OOD detection performance in terms of AUROC, AUPR-Success, AUPR-Error, and FPR at 95\% TPR. The results are averaged over the number of OOD test sets. We compare our approach to methods that do not further optimize the classifier but operate on the pre-trained model. The results marked with $^{\ast}$ are taken from the original paper. $\uparrow$ indicates that larger values are better, whereas $\downarrow$ marks that lower values are better.}
    \label{tab:ood_results}
\end{table*}

\subsection{Out-of-Distribution Detection Results}\label{sec:oodresults}
The OOD detection results are presented in Tab.~\ref{tab:ood_results}. Similar to the related work~\cite{hendrycks2018deep,liu20_energy}, we report the performance averaged over the respective OOD test sets. A detailed evaluation of the individual OOD datasets is provided in the supplementary material. 
In general, the degree of difficulty increases with a larger class number and higher resolution. Thus, the OOD detection for CIFAR-10 is less complex, while for CIFAR-100 it is more difficult and Tiny ImageNet is the most complex setup in our experiments. 
In the case of CIFAR-10, the built-in softmax score already achieves a good OOD detection performance in terms of AUROC and AUPR-S.
The other related approaches besides ReAct improve the MSP especially in the detection of OOD samples (AUPR-E) and the FPR-95 metric. Nevertheless, we obtain better results in all cases, especially for the FPR-95 metric.
For CIFAR-100, the OOD detection performance of all methods decreases with $100$ instead of $10$ in-distribution classes. Here, we outperform the other approaches in all metrics. For both architectures, the AUPR-E metric highlights that the OOD detection, in particular, is significantly improved compared to the other methods.
The last experiment, with Tiny ImageNet as an in-distribution dataset, emphasizes that the more classes that need to be categorized, the harder it becomes to detect OOD samples. In this setup, we achieve comparable or even better results (AUPR-E, FPR-95) than ReAct and outperform the other methods in all metrics. 
Overall, the metrics indicate that all methods have a tendency towards the detection of in-distribution samples. This is shown by the higher value for AUPR-S in contrast to AUPR-E. However, our approach surpasses prior work in the detection of OOD samples with an improvement of almost 10\% in terms of AUPR-E for the majority of the cases and works consistently for all cases.
\subsection{Visual Illustrations}\label{sec:visu}
In Fig.~\ref{fig:examples}, we demonstrate example heatmap definitions used for the decoder training. Fig.~\ref{fig:example_ood} shows OOD images from Places365, Fig.~\ref{fig:example_id} visualizes their closest in-distribution training sample from Tiny ImageNet, while in Fig.~\ref{fig:example_hm} the resulting heatmap is illustrated. In Fig.~\ref{fig:examples}, the blue colors mark similar regions, whereas the red/yellow colors highlight features that differ from the in-distribution image. The first row represents a far-distribution image, while the second row represents a near-distribution image. Here, the near-distribution heatmap shows a milder response compared to the far-distribution heatmap.
\begin{figure}[ht]
    \centering
    \begin{tabular}{@{}c@{\hskip2pt}|@{\hskip1pt}@{\hskip1pt}c@{\hskip1pt}@{\hskip1pt}c@{\hskip1pt}@{\hskip1pt}c@{\hskip1pt}}
    \textbf{In} & \multicolumn{3}{c}{\textbf{Out}} \\
    \includegraphics[width=0.2\linewidth]{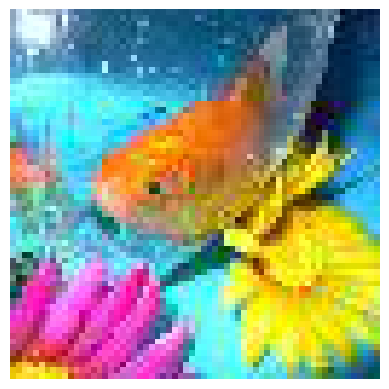}
    &
    \includegraphics[width=0.2\linewidth]{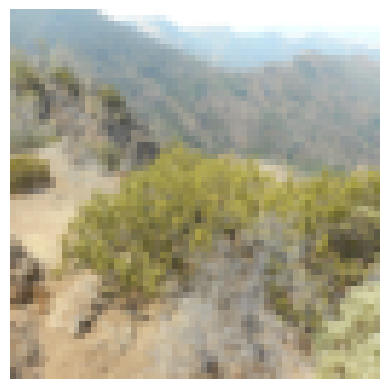}
    & 
    \includegraphics[width=0.2\linewidth]{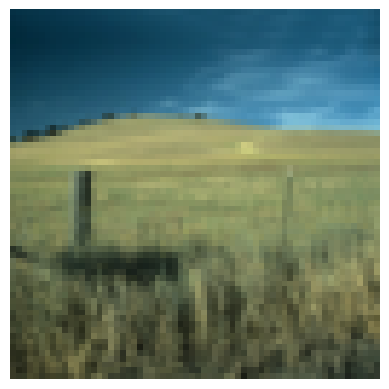}
    & 
    \includegraphics[width=0.2\linewidth]{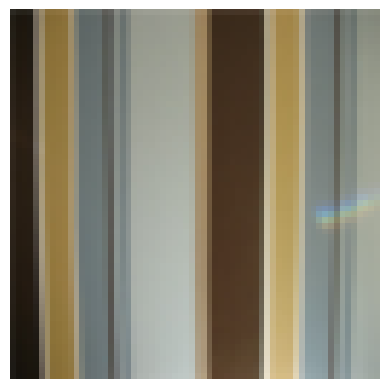}
    \\
    \subfloat[\label{fig:ti_a}]{\includegraphics[width=0.2\linewidth]{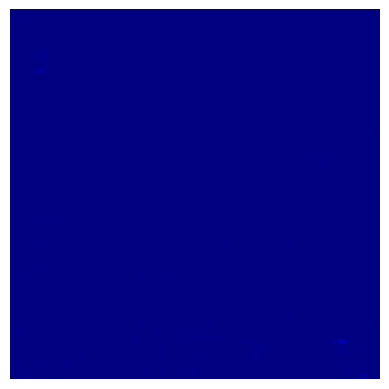}}
    & 
    \subfloat[\label{fig:ti_b}]{\includegraphics[width=0.2\linewidth]{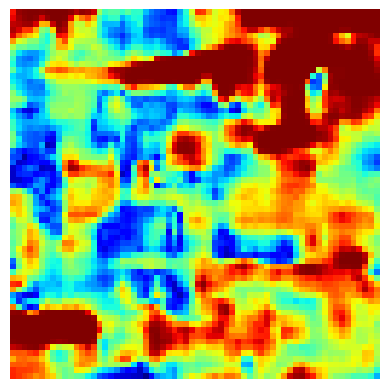}}
    & 
    \subfloat[\label{fig:ti_c}]{\includegraphics[width=0.2\linewidth]{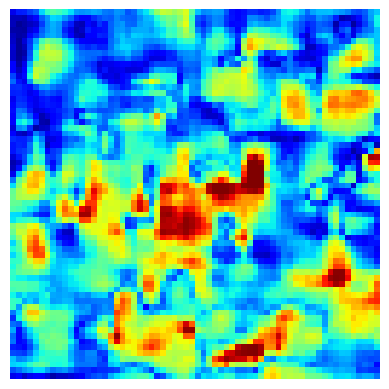}}
    & 
    \subfloat[\label{fig:ti_d}]{\includegraphics[width=0.2\linewidth]{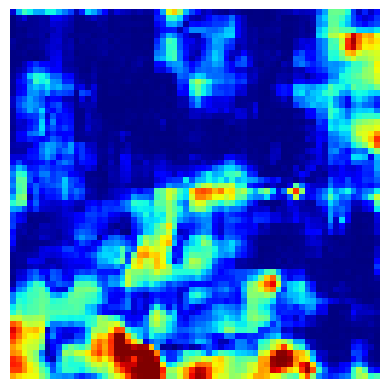}} \\
    \end{tabular}
    \caption{Visual results with ResNet trained on TinyImageNet (a) as an in-distribution database (\textbf{In}). Examples from the OOD databases (\textbf{Out}) iNaturalist (b), SUN  (c) and Textures (d) are shown. The original images are displayed in the top row, whereas the estimated heatmaps are in the bottom row. Blue colors mark in-distribution regions, whereas the red/yellow colors highlight OOD regions.} 
    \label{fig:visu_imgnet}
\end{figure}

Fig.~\ref{fig:visu_imgnet} shows qualitative results for ResNet trained on Tiny ImageNet. The first row presents the original input images, while the estimated heatmaps are in the second row. Fig.~\ref{fig:ti_a} visualizes an in-distribution example of Tiny ImageNet. The heatmap shows no response compared to the original image. Thus, the heatmap entries are close to zero, which in turn means that features extracted with the classifier are representative of in-distribution samples. From Fig.~\ref{fig:ti_b} to Fig.~\ref{fig:ti_d}, we illustrate three different OOD examples as input to the classifier trained on Tiny ImageNet. In all cases, regions over the entire images are highlighted by out-of-distribution responses indicating the differences from the closest in-distribution sample. The OOD examples cover different OOD types with landscapes (Fig.~\ref{fig:ti_b}, Fig.~\ref{fig:ti_c}) and textures (Fig.~\ref{fig:ti_d}) demonstrating the generalization capability of our approach. Further visual results are provided in the supplementary material.

\subsection{Lighting Effect}\label{sec:br}
\begin{figure}[ht]
    \centering
    \begin{tabular}{c@{\hskip1pt}@{\hskip1pt}c@{\hskip0pt}@{\hskip0pt}c@{\hskip1pt}@{\hskip1pt}c@{\hskip0pt}@{\hskip0pt}c}
    \includegraphics[width=0.19\linewidth]{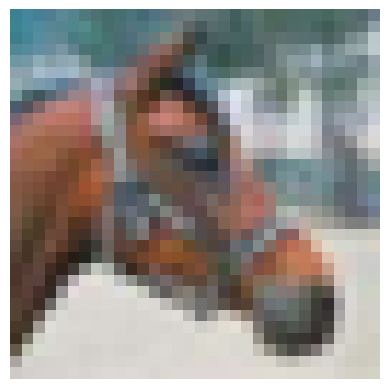}
    &
    \includegraphics[width=0.19\linewidth]{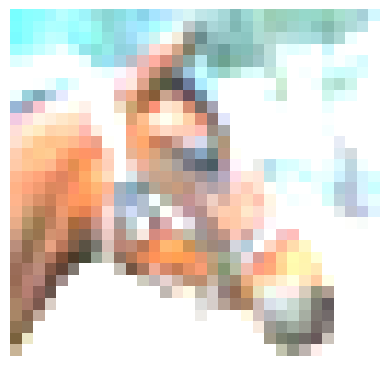}
    &
    \includegraphics[width=0.19\linewidth]{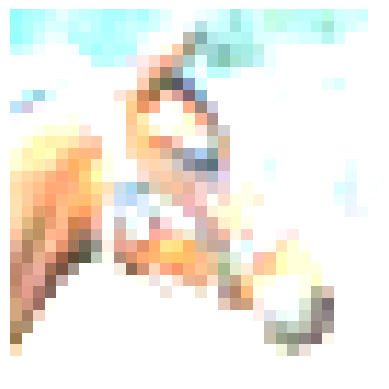}
    &
    \includegraphics[width=0.19\linewidth]{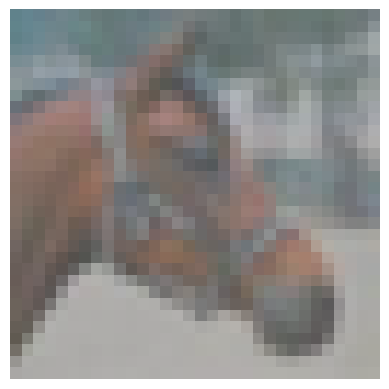}
    &
    \includegraphics[width=0.19\linewidth]{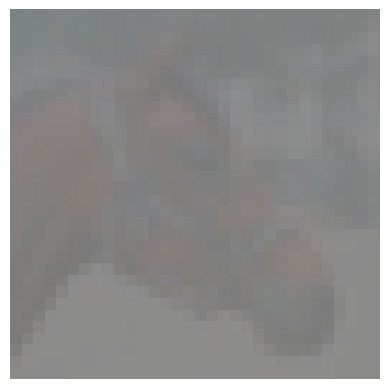}
    \\
    \subfloat[\label{fig:l_a}]{\includegraphics[width=0.19\linewidth]{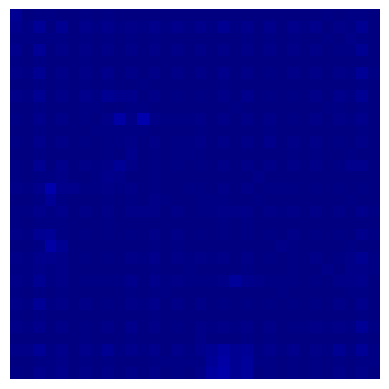}}
    &
    \subfloat[\label{fig:l_b}B=2.0]{\includegraphics[width=0.19\linewidth]{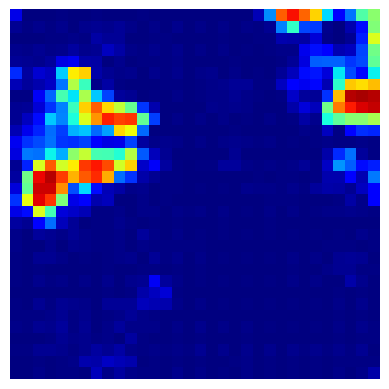}}
    &
    \subfloat[\label{fig:l_c}B=2.5]{\includegraphics[width=0.19\linewidth]{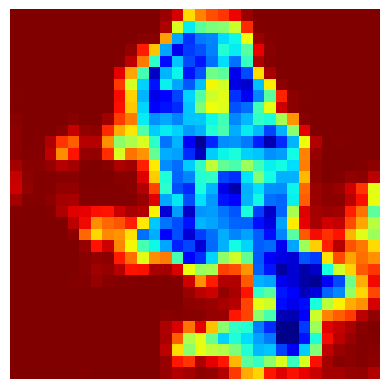}}
    &
    \subfloat[\label{fig:l_d}C=0.5]{\includegraphics[width=0.19\linewidth]{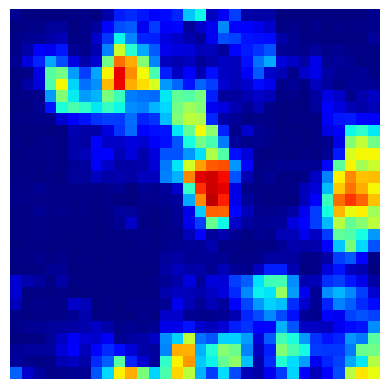}}
    &
    \subfloat[\label{fig:l_e}C=0.1]{\includegraphics[width=0.19\linewidth]{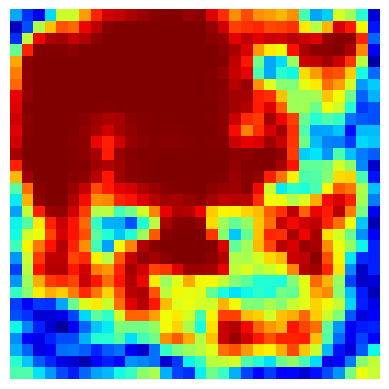}}
    \\
    \end{tabular}
    \caption{Example images of a horse from the CIFAR-10 testset with corresponding heatmap predictions augmented with different brightness (B) or contrast (C) values. (a) is the original image without augmentation, whereas in (b) and (c) the brightness value is increased. In (d) and (e), the contrast is decreased. Blue colors mark in-distribution regions, whereas the red/yellow colors highlight out-of-distribution regions.}
    \label{fig:lighting}
\end{figure}
The performance of neural networks can be affected by external factors such as lighting conditions. Therefore, we evaluate the influence of brightness and contrast changes on our approach. We augment in-distribution test data with increased brightness (B) selected from $\text{B} = \{2.0, 2.5\}$ and reduced contrast (C) selected from $\text{C} = \{0.1, 0.5\}$. The higher the brightness and the lower the contrast, the more augmented in-distribution samples should be detected as OOD, since the relevant features are no longer recognizable. In Fig.~\ref{fig:lighting}, examples of a horse with augmented brightness and augmented contrast are visualized with the corresponding heatmap predictions. The heatmaps clearly show a higher response for images with higher brightness augmentation as well as for images with lower contrast augmentation. With increasing brightness the in-distribution features of the images are less recognizable and therefore the images should be labelled as OOD. The same applies to images with reduced contrast. In general, more image regions are visualized as OOD for increasing brightness values. For $\text{B}=2.5$ (Fig.~\ref{fig:l_c}), the pixels covering the horse's head are still marked as in-distribution, while most other pixels are highlighted as OOD. By contrast, when the brightness and contrast are not changed (Fig.~\ref{fig:l_a}), the heatmap entries are zero. For the image augmented with $\text{C}=0.1$ (Fig.~\ref{fig:l_e}), the corresponding heatmap also shows larger OOD regions compared to the original image. In the supplementary material, we report the OOD detection performance evaluation where the augmented in-distribution samples are labelled as OOD. As already evident from the qualitative heatmap examples, the OOD detection performance increases with higher brightness and reduced contrast. 

\subsection{Discussion}
Overall, we implement a trainable model to generate the heatmaps. Since we optimize the parametrized model, we naturally have additional computational effort in comparison to prior work~\cite{hendrycks17_baseline,Liang2018EnhancingTR}.
Nevertheless, the heatmaps can not only be leveraged to detect OOD samples but also as a possibility to illustrate in- and out-of-distribution image regions based on the classifier features and the class prediction.
At the moment, our approach is specifically designed for image classification models. Since the reference images are associated with the class prediction, the approach is limited to classification tasks. The adaptation to other domains, such as object detection, could be addressed in future work.  

\subsection{\textbf{Ablation Study}}\label{sec:ablation}
\begin{figure}
    \centering
    \includegraphics[width=0.95\linewidth]{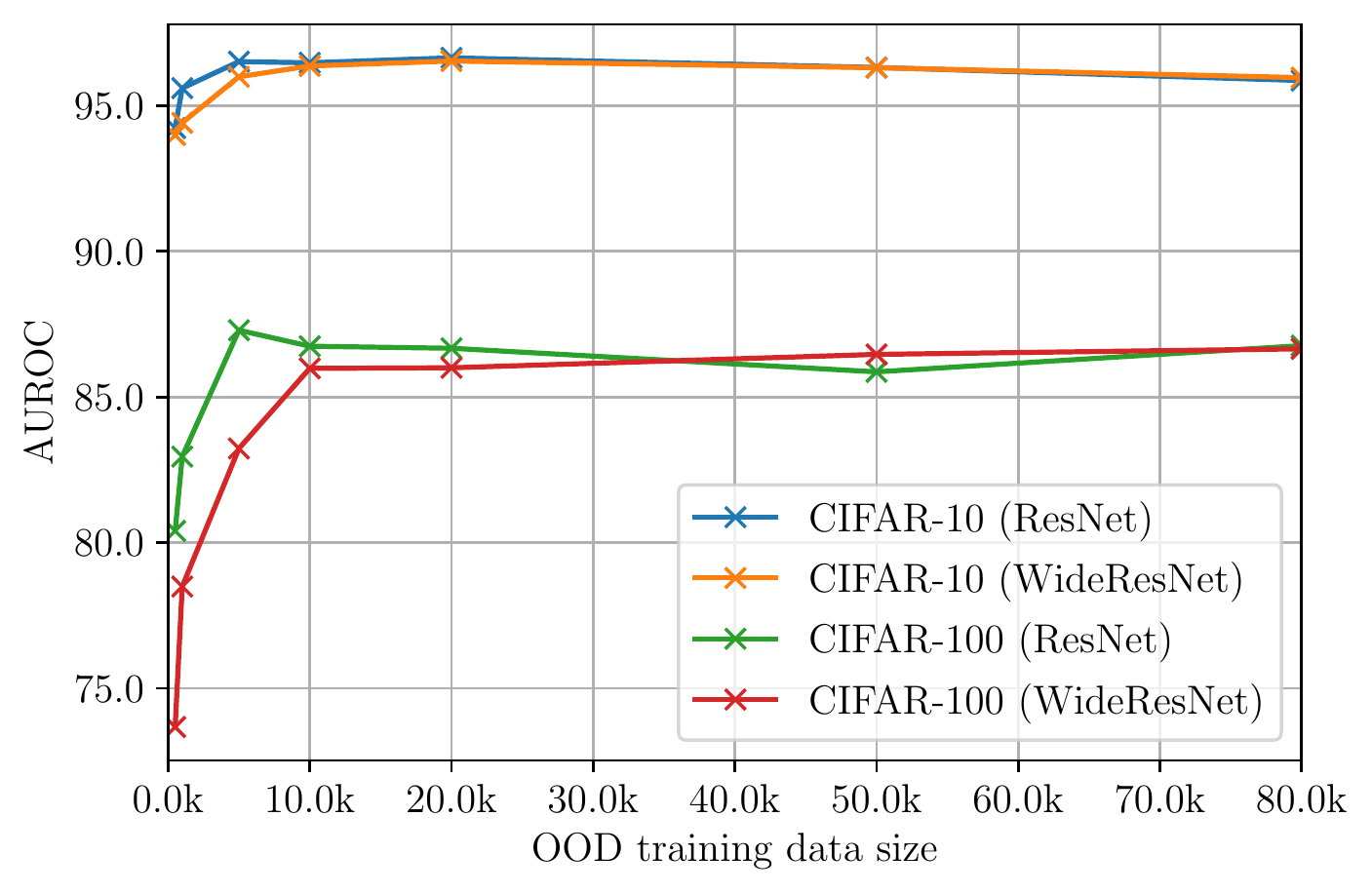}
    \caption{AUROC results when alternating the OOD training set. The in-distribution dataset size is fixed.}
    \label{fig:nsamples}
\end{figure}

We study the influence of the OOD training data size. Since AUPR-S and AUPR-E set the focus on the positive class, we report the AUROC and provide the other metrics in the supplementary material. 
We conduct the ablation studies with both network architectures pre-trained on CIFAR-10 and CIFAR-100 as in-distribution dataset and 80 Million Tiny Images as OOD training set. As mentioned in Sec.~\ref{sec:implementation}, the OOD training set size is set to $\frac{1}{5}$ of the in-distribution dataset size. We keep the number of in-distribution samples fixed at $50K$ and vary the number of OOD samples. In Fig.~\ref{fig:nsamples}, we present the AUROC results when alternating the OOD training set size with sets of $\{500, 1000, 5000, 10000, 20000, 50000, 80000\}$ samples.
Especially for the WideResNet architecture, the performance considerably drops when using less than $10K$ samples. Between $10K$ to $80K$ OOD samples, the AUROC slightly decreases for both datasets and both architectures. The minor deterioration in performance can be explained by the increasing focus on OOD samples. Since the performance degradation is negligible, the OOD training set size should be at least $\frac{1}{5}$ of the in-distribution dataset size.

\section{Conclusion}
\label{sec:conclusion}
We presented an approach to learn heatmaps for OOD detection, which also serve as an illustration of in- and out-of-distribution image regions.
We introduced the decoder that is trained to produce heatmaps zero response for an in-distribution sample and high response for OOD samples, based on the classifier features and the class prediction of a fixed classifier. Our main contribution consists of the OOD sample heatmap definition that is based on the normalized difference from the closest in-distribution sample. The heatmap eventually acts as a margin to distinguish between in- and out-of-distribution images. In our evaluations, we show that our OOD score function, based on the heatmap response, achieves state-of-the-art OOD detection performance compared to fixed classifiers approaches, trained on CIFAR-10, CIFAR-100, and Tiny ImageNet.

\paragraph{\textbf{Acknowledgements.}} The research leading to these results is funded by the German Federal Ministry for Economic Affairs and Climate Action" within the project “KI Delta Learning“ (Förderkennzeichen 19A19013A). The authors would like to thank the consortium for the successful cooperation.

{\small
\bibliographystyle{ieee_fullname}
\bibliography{wacv_2023}
}

\end{document}